\documentclass{article} % For LaTeX2e
\usepackage{iclr2016_conference,times}
\usepackage{hyperref}
\usepackage{url}
\usepackage[pdftex]{graphicx}
\usepackage{amsmath}
\usepackage{array}
\usepackage{algorithm}
\usepackage[noend]{algpseudocode}

\title{Neural Networks with Few Multiplications}

\author{Zhouhan Lin \\  %\thanks{ Use footnote for providing further information about author (webpage, alternative address)---\emph{not} for acknowledging funding agencies.} \\
Universit\'e de Montr\'eal\\
Canada \\
\texttt{zhouhan.lin@umontreal.ca} \\
\And
Matthieu Courbariaux \\
Universit\'e de Montr\'eal\\
Canada \\
\texttt{matthieu.courbariaux@gmail.com} \\
\AND
Roland Memisevic \\
Universit\'e de Montr\'eal\\
Canada \\
\texttt{roland.umontreal@gmail.com} \\
\And
Yoshua Bengio \\
Universit\'e de Montr\'eal\\
Canada}

\iclrfinalcopy % Uncomment for camera-ready version

\begin{document}

\maketitle

\begin{abstract}
For most deep learning algorithms training is notoriously time consuming. 
Since most of the computation in training neural networks is typically spent on floating point
multiplications, we investigate an approach to training that eliminates the need for most of these.  
%That also makes it possible to train a network very fast on a properly designed chip. 
Our method consists of two parts: 
First we stochastically binarize weights to convert multiplications involved in computing hidden states to sign changes. Second, while back-propagating error derivatives, in addition to binarizing the weights, we quantize the representations at each layer to convert the remaining multiplications into binary shifts.
Experimental results across 3 popular datasets (MNIST, CIFAR10, SVHN) show that this approach not only does not hurt classification performance but can result in even better performance than standard stochastic gradient descent training, paving the way to fast, hardware-friendly training of neural networks. 
\end{abstract}

\section{Introduction}
Training deep neural networks has long been computational demanding and time
consuming. For some state-of-the-art architectures, it can take weeks to
get models trained \citep{krizhevsky2012}. Another problem is that the demand for memory can be 
huge. For example, many common models in speech recognition or machine
translation need 12 Gigabytes or more of storage \citep{gulcehre2015using}. 
%These two demands in both memory and computation make training a 
To deal with these issues it is common to train 
deep neural networks by resorting to GPU or CPU clusters and to well designed parallelization strategies
\citep{le2013building}. 

Most of the computation performed in training a neural network are floating
point multiplications. In this paper, we focus on eliminating most of these multiplications to reduce computation. Based on our previous work \citep{binaryconnect}, which eliminates multiplications in computing hidden representations by binarizing weights, our method deals with both hidden state computations and backward weight updates. Our approach has 2 components. In the forward pass, weights are stochastically binarized using an approach we call \textit{binary connect} or \textit{ternary connect}, and for back-propagation of errors, we propose a new approach which we call \textit{quantized back
propagation} that converts multiplications into bit-shifts. \footnote{The codes for these approaches are available online at \url{https://github.com/hantek/BinaryConnect}}

\section{Related work}
Several approaches have been proposed in the past to simplify computations in neural networks. Some of them try to restrict weight values to be an integer power of two, thus to reduce all the multiplications to be binary shifts \citep{kwan1993multiplierless, marchesi1993fast}. In this way, multiplications are eliminated in both training and testing time. The disadvantage is that model performance can be severely reduced, and convergence of training can no longer be guaranteed. 

\cite{bitwisenet} introduces a completely Boolean network, which simplifies the test time computation at an acceptable performance hit. The approach still requires a real-valued, full precision training phase, however, so the benefits of reducing computations does not apply to training. Similarly, \cite{machado2015computational} manage to get acceptable accuracy on sparse representation classification by replacing all floating-point multiplications by integer shifts. Bit-stream networks \citep{Burge:1999:SBN:296533.296552} also provides a way of binarizing neural network connections, by substituting weight connections with logical gates. Similar to that, \cite{cheng2015training} proves deep neural networks with binary weights can be trained to distinguish between multiple classes with expectation back propagation.

There are some other techniques, which focus on reducing the training complexity. For instance, instead of reducing the precision of weights, \cite{simard1994backpropagation} quantizes states, learning rates, and gradients to powers of two. This approach manages to eliminate multiplications with negligible performance reduction.

\section{Binary and ternary connect}
\subsection{Binary connect revisited}
In \cite{binaryconnect}, we introduced a weight binarization technique which removes multiplications in the forward pass. We summarize this approach in this subsection, and introduce an extension to it in the next. 

Consider a neural network layer with $N$ input and $M$ output units. The forward computation is $\mathbf{y} = h(W\mathbf{x+b})$ where $W$ and $\mathbf{b}$ are weights and biases, respectively, $h$ is the activation function, and $\mathbf{x}$ and $\mathbf{y}$ are the layer's inputs and outputs. If we choose ReLU as $h$, there will be no multiplications in computing the activation function, thus all multiplications reside in the matrix product $W\mathbf{x}$. For each input vector $x$, $NM$ floating point multiplications are needed.

Binary connect eliminates these multiplications by stochastically sampling weights to be $-1$ or $1$. Full precision weights $\bar{w}$ are kept in memory as reference, and each time when $y$ is needed, we sample a stochastic weight matrix $W$ according to $\bar{w}$. For each element of the sampled matrix $W$, the probability of getting a $1$ is proportional to how ``close" its corresponding entry in $\bar{w}$ is to $1$. i.e., 

\begin{equation}  \label{pwij}
P(W_{ij}=1)=\frac{\bar{w}_{ij}+1}{2}; \; \; P(W_{ij}=-1)=1-P(W_{ij}=1)
\end{equation}

It is necessary to add some edge constraints to $\bar{w}$. To ensure that $P(W_{ij}=1)$ lies in a reasonable range, values in $\bar{w}$ are forced to be a real value in the interval [-1, 1]. If during the updates any of its value grows beyond that interval, we set it to be its corresponding edge values $-1$ or $1$. That way floating point multiplications become sign changes.

A remaining question concerns the use of multiplications in the random number generator involved in the sampling process. Sampling an integer has to be faster than multiplication for the algorithm to be worth it. To be precise, in most cases we are doing mini-batch learning and the sampling process is performed only once for the whole mini-batch. Normally the batch size $B$ varies up to several hundreds. So, as long as one sampling process is significantly faster than $B$ times of multiplications, it is still worth it. Fortunately, efficiently generating random numbers has been studied in \cite{jeavons1994generating, van1993device}. Also, it is possible to get random numbers according to real random processes, like CPU temperatures, etc. We are not going into the details of random number generation as this is not the focus of this paper.

\subsection{Ternary connect}
The binary connect introduced in the former subsection allows weights to be $-1$ or $1$. However, in a trained neural network, it is common to observe that many learned weights are zero or close to zero. 
%If no zeros are allowed in the sampled weight values, that probably means in all of the time we are forcing the weights sampled to be a value far away from its actual values. 
Although the stochastic sampling process would allow the mean value of sampled weights to be zero, this suggests that it may be beneficial to explicitly allow weights to be zero. 

To allow weights to be zero, some adjustments are needed for Eq. \ref{pwij}. We split the interval of [-1, 1], within which the full precision weight value $\bar{w_{ij}}$ lies, into two sub-intervals: [$-1, 0$] and ($0, 1$]. If a weight value $\bar{w}_{ij}$ drops into one of them, we sample $\bar{w}_{ij}$ to be the two edge values of that interval, according to their distance from $\bar{w}_{ij}$, i.e., if $\bar{w}_{ij}>0$:

\begin{equation}
P(W_{ij}=1) = \bar{w}_{ij}; \; \;P(W_{ij}=0) = 1-\bar{w}_{ij}
\end{equation}

and if $\bar{w_{ij}}<=0$: 

\begin{equation}
P(W_{ij}=-1) = -\bar{w}_{ij}; \; \; P(W_{ij}=0) = 1+\bar{w}_{ij}
\end{equation}

Like binary connect, ternary connect also eliminates all multiplications in the forward pass.

%The way the network binarize the weights are:
%1. calculate the absolute value of each weight value, if it is larger than 1, clip it to be 1.
%2. sample a Bernoulli distribution, according to the absolute value calculated in ``step 1". (The closer the absolute weight values are to ``1", the more likely a ``1" can be sampled.)
%3. add the sign back, so there are three possible values {0, -1, 1} in the weight matrix.

\section{Quantized back propagation} \label{qbp}
In the former section we described how multiplications can be eliminated from the forward pass. 
%However, it also involves multiplications while propagating error signal through the network during backward pass. 
In this section, we propose a way to eliminate multiplications from the backward pass.

Suppose the $i$-th layer of the network has $N$ input and $M$ output units, and consider an error signal $\mathbf{\delta}$ propagating downward from its output. The updates for weights and biases would be the outer product of the layer's input and the error signal:

\begin{equation} \label{deltaw}
\Delta W = \eta \left[ \mathbf{\delta} \odot h^{'}\left( W\mathbf{x} + b \right) \right] \mathbf{x}^T
\end{equation}
\begin{equation} \label{deltab}
\Delta b = \eta \left[ \mathbf{\delta} \odot h^{'}\left( W\mathbf{x} + b \right) \right]
\end{equation}

where $\eta$ is the learning rate, and $\mathbf{x}$ the input to the layer. The operator $\odot$ stands for element-wise multiply. While propagating through the layers, the error signal $\mathbf{\delta}$ needs to be updated, too. Its update taking into account the next layer below takes the form:

\begin{equation} \label{newdelta}
\mathbf{\delta} = \left[ W^T \mathbf{\delta} \right] \odot h^{'}\left( W\mathbf{x} + b \right)
\end{equation}

There are 3 terms that appear repeatedly in Eqs. \ref{deltaw} to \ref{newdelta}: $\mathbf{\delta}, h^{'}\left(W\mathbf{x} + b \right)$ and $\mathbf{x}$. The latter two terms introduce matrix outer products. To eliminate multiplications, we can quantize one of them to be an integer power of $2$, so that multiplications involving that term become binary shifts. The expression $h^{'}\left( W\mathbf{x} + b \right)$ contains downflowing gradients, which are largely determined by the cost function and network parameters, thus it is hard to bound its values. However, bounding the values is essential for quantization because we need to supply a fixed number of bits for each sampled value, and if that value varies too much, we will need too many bits for the exponent. This, in turn, will result in the need for more bits to store the sampled value and unnecessarily increase the required amount of computation. 

While $h^{'}\left( W\mathbf{x} + b \right)$ is not a good choice for quantization, $\mathbf{x}$ is a better choice, because it is the hidden representation at each layer, and we know roughly the distribution of each layer's activation. 

Our approach is therefore to eliminate multiplications in Eq. \ref{deltaw} by quantizing each entry in $\mathbf{x}$ to an integer power of $2$. That way the outer product in Eq. \ref{deltaw} becomes a series of bit shifts. Experimentally, we find that allowing a maximum of $3$ to $4$ bits of shift is sufficient to make the network work well. This means that $3$ bits are already enough to quantize $\mathbf{x}$. As the float32 format has $24$ bits of mantissa, shifting (to the left or right) by $3$ to $4$ bits is completely tolerable. We refer to this approach of back propagation as ``quantized back propagation."

If we choose ReLU as the activation function, and since we are reusing the $\left( W\mathbf{x} + b \right)$ that was computed during the forward pass, computing the term $h^{'}\left( W \mathbf{x} + b \right)$ involves no additional sampling or multiplications. In addition, quantized back propagation eliminates the multiplications in the outer product in Eq. \ref{deltaw}. The only places where multiplications remain are the element-wise products. In Eq. \ref{deltab}, multiplying by $\eta$ and $\sigma$ requires $2\times M$ multiplications, while in Eq. \ref{deltaw} we can reuse the result of Eq. \ref{deltab}. To update $\delta$ would need another $M$ multiplications, thus $3\times M$ multiplications are needed for all computations from Eqs. \ref{deltaw} through \ref{newdelta}. Pseudo code in Algorithm \ref{qbpalgo} outlines how quantized back propagation is conducted.

\begin{algorithm}[H]
\caption{Quantized Back Propagation (QBP). $C$ is the cost function. ${\rm binarize}(W)$ and ${\rm clip}(W)$ stands for binarize and clip methods. $L$ is the number of layers.}
\label{qbpalgo}
\begin{algorithmic}[1]
    \Require a deep model with parameters $W$, $b$ at each layer. 
             Input data $x$, its corresponding targets $y$, and learning rate $\eta$.
    \Procedure{QBP}{model, $x$, $y$, $\eta$}
    
    \State {\bf 1. Forward propagation:}
    \For{each layer $i$ in ${\rm range}(1, L)$}
    \State     $W_b \leftarrow {\rm binarize}(W)$
    \State     Compute activation $a_i$ according to its 
               previous layer output $a_{i-1}$, $W_b$ and $b$.
    \EndFor
    
    \State {\bf 2. Backward propagation:}
    \State Initialize output layer's error signal $\delta=\frac{\partial C}{\partial a_L}$.
    \For{each layer $i$ in ${\rm range}(L, 1)$}
    \State Compute $\Delta W$ and $\Delta b$ according to Eqs. \ref{deltaw} and \ref{deltab}.
    \State Update $W$: $W \leftarrow {\rm clip}(W - \Delta W)$
    \State Update $b$: $b \leftarrow b - \Delta b$
    \State Compute $\frac{\partial C}{\partial a_{k-1}}$ by updating $\delta$
           according to Eq. \ref{newdelta}.
    \EndFor
    
    \EndProcedure
\end{algorithmic}
\end{algorithm}

Like in the forward pass, most of the multiplications are used in the weight updates. Compared with standard back propagation, which would need $2MN + 3M$ multiplications at least, the amount of multiplications left is negligible in quantized back propagation. Our experiments in Section \ref{exp} show that this way of dramatically decreasing multiplications does not necessarily entail a loss in performance.

\section{Experiments} \label{exp}
We tried our approach on both fully connected networks and convolutional networks. Our implementation uses Theano \citep{Bastien-Theano-2012}. We experimented with $3$ datasets: MNIST, CIFAR10, and SVHN. In the following subsection we show the performance that these multiplier-light neural networks can achieve. In the subsequent subsections we study some of their properties, such as convergence and robustness, in more detail. 

\subsection{General performance}
We tested different variations of our approach, and compare the results with \cite{binaryconnect} and full precision training (Table \ref{performance}). All models are trained with stochastic gradient descent (SGD) without momentum. We use batch normalization for all the models to accelerate learning. At training time, binary (ternary) connect and quantized back propagation are used, while at test time, we use the learned full resolution weights for the forward propagation. For each dataset, all hyper-parameters are set to the same values for the different methods, except that the learning rate is adapted independently for each one.

\begin{table}[!h]
    \caption{Performances across different datasets}  \label{performance}
    \begin{center}
        \begin{tabular}{c | c | c | c | c}
        \hline\hline
	        & Full precision	& Binary connect
	        & \begin{tabular}{c} Binary connect + \\ Quantized backprop
	          \end{tabular}
	        & \begin{tabular}{c} Ternary connect + \\ Quantized backprop
	          \end{tabular}  \\
        \hline
        MNIST	& 1.33\%	& 1.23\%    & 1.29\%    & 1.15\%    \\
        CIFAR10	& 15.64\%	& 12.04\%	& 12.08\%	& 12.01\%   \\
        SVHN	& 2.85\%    & 2.47\%    & 2.48\%    & 2.42\%    \\
        \hline\hline
        \end{tabular}
    \end{center}
\end{table}

\subsubsection{MNIST}
The MNIST dataset \citep{lecun1998gradient} has 50000 images for training and 10000 for testing. All images are grey value images of size $28\times28$ pixels, falling into 10 classes corresponding to the 10 digits. The model we use is a fully connected network with 4 layers: 784-1024-1024-1024-10. At the last layer we use the hinge loss as the cost. The training set is separated into two parts, one of which is the training set with 40000 images and the other the validation set with 10000 images. Training is conducted in a mini-batch way, with a batch size of 200. 

With ternary connect, quantized backprop, and batch normalization, we reach an error rate of 1.15\%. This result is better than full precision training (also with batch normalization), which yields an error rate 1.33\%. If without batch normalization, the error rates rise to 1.48\% and 1.67\%, respectively. We also explored the performance if we sample those weights during \textit{test time}. With ternary connect at test time, the same model (the one reaches 1.15\% error rate) yields 1.49\% error rate, which is still fairly acceptable. Our experimental results show that despite removing most multiplications, our approach yields a comparable (in fact, even slightly higher) performance than full precision training. The performance improvement is likely due to the regularization effect implied by the stochastic sampling. 

Taking this network as a concrete example, the actual amount of multiplications in each case can be estimated precisely. Multiplications in the forward pass is obvious, and for the backward pass section \ref{qbp} has already given an estimation. Now we estimate the amount of multiplications incurred by batch normalization. Suppose we have a pre-hidden representation $h$ with mini-batch size $B$ on a layer which has $M$ output units (thus $h$ should have shape $B \times M$), then batch normalization can be formalized as $\gamma \frac{h - mean(h)}{std(h)} + \beta$. One need to compute the $mean(h)$ over a mini-batch, which takes $M$ multiplications, and $BM+2M$ multiplication to compute the standard deviation $std(h)$. The fraction takes $BM$ divisions, which should be equal to the same amount of multiplication. Multiplying that by the $\gamma$ parameter, adds another $BM$ multiplications. So each batch normalization layer takes an extra $3BM+3M$ multiplications in the forward pass. The backward pass takes roughly twice as many multiplications in addition, if we use SGD. These amount of multiplications are the same no matter we use binarization or not. Bearing those in mind, the total amount of multiplications invoked in a mini-batch update are shown in Table \ref{multiplications}. The last column lists the ratio of multiplications left, after applying ternary connect and quantized back propagation.

\begin{table}[!h]
    \caption{Estimated number of multiplications in MNIST net}  \label{multiplications}
    \begin{center}
        \begin{tabular}{c | c | c | c}
        \hline\hline
	        & Full precision
	        & \begin{tabular}{c} Ternary connect + \\ Quantized backprop
	          \end{tabular}
	        & ratio \\
        \hline
        without BN	& $1.7480\times10^9$	& $1.8492\times10^6$    & 0.001058    \\
        with BN	    & $1.7535\times10^9$	& $7.4245\times10^6$	& 0.004234   \\
        \hline\hline
        \end{tabular}
    \end{center}
\end{table}

% the computation behind the above table is (B=200 for the settings in the paper):
%  3BMN + 3BM   |   3BM
%  3BMN+12BM+9M |   12BM+9M
% It could be derived by the amounts written in the paper. i.e.
%  (3MN+3M)*B              |   3M*B
%  (3MN+3M)*B + 3(3BM+3M)  |   3M*B + 3(3BM+3M)

\subsubsection{CIFAR10}
CIFAR10 \citep{krizhevsky2009learning} contains images of size $32\times32$ RGB pixels. Like for MNIST, we split the dataset into 40000, 10000, and 10000 training-, validation-, and test-cases, respectively. We apply our approach in a convolutional network for this dataset. The network has 6 convolution/pooling layers, 1 fully connected layer and 1 classification layer. We use the hinge loss for training, with a batch size of 100. We also tried using ternary connect at test time. On the model trained by ternary connect and quantized back propagation, it yields 13.54\% error rate. Similar to what we observed in the fully connected network, binary (ternary) connect and quantized back propagation yield a slightly higher performance than ordinary SGD. 

\subsubsection{SVHN}
The Street View House Numbers (SVHN) dataset \citep{netzer2011reading} contains RGB images of house numbers. It contains more than 600,000 images in its extended training set, and roughly 26,000 images in its test set. We remove 6,000 images from the training set for validation. We use 7 layers of convolution/pooling, 1 fully connected layer, and 1 classification layer. Batch size is also set to be 100. The performances we get is consistent with our results on CIFAR10. Extending the ternary connect mechanism to its test time yields 2.99\% error rate on this dataset. Again, it improves over ordinary SGD by using binary (ternary) connect and quantized back propagation. 

\subsection{Convergence}
\begin{figure}[!h]
\begin{center}
\includegraphics[width=0.7\linewidth]{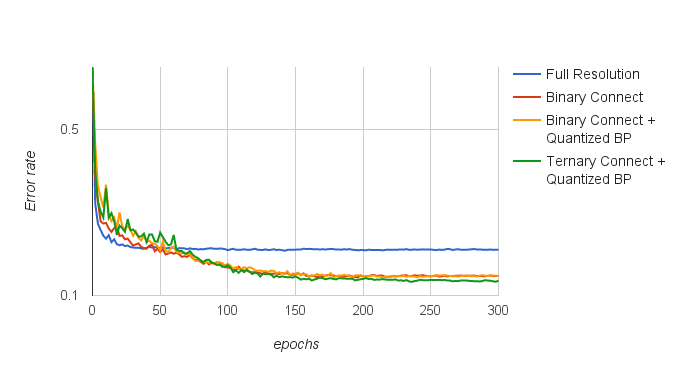}
\caption{Test set error rate at each epoch for ordinary back propagation, binary connect, binary connect with quantized back propagation, and ternary connect with quantized back propagation. Vertical axis is represented in logarithmic scale.}
\label{epc}
\end{center}
\end{figure}

Taking the convolutional networks on CIFAR10 as a test-bed, we now study the learning behaviour in more detail. Figure \ref{epc} shows the performance of the model in terms of test set errors during training. The figure shows that binarization makes the network converge slower than ordinary SGD, but yields a better optimum after the algorithm converges. Compared with binary connect (red line), adding quantization in the error propagation (yellow line) doesn't hurt the model accuracy at all. Moreover, having ternary connect combined with quantized back propagation (green line) surpasses all the other three approaches.

\subsection{The effect of bit clipping}
\begin{figure}[!h]
\begin{center}
\includegraphics[width=0.63\linewidth]{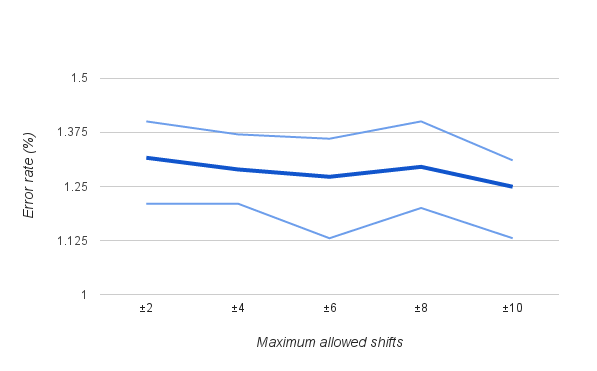}
\caption{Model performance as a function of the maximum bit shifts allowed in quantized back propagation. The dark blue line indicates mean error rate over 10 independent runs, while light blue lines indicate their corresponding maximum and minimum error rates. %The numbers in horizontal line indicate maximum number of bits allowed to be shifted (to the left or right), while vertical line corresponds to the error rate yielded.
}
\label{qbplimit}
\end{center}
\end{figure}

In Section \ref{qbp} we mentioned that quantization will be limited by the number of bits we use. The maximum number of bits to shift determines the amount of memory needed, but it also determines in what range a single weight update can vary. Figure~\ref{qbplimit} shows the model performance as a function of the maximum allowed bit shifts. These experiments are conducted on the MNIST dataset, with the aforementioned fully connected model. For each case of bit clipping, we repeat the experiment for 10 times with different initial random instantiations. 

The figure shows that the approach is not very sensible to the number of bits used. The maximum allowed shift in the figure varies from 2 bits to 10 bits, and the performance remains roughly the same. Even by restricting bit shifts to 2, the model can still learn successfully. The fact that the performance is not very sensitive to the maximum of allowed bit shifts suggests that we do not need to redefine the number of bits used for quantizing $\mathbf{x}$ for different tasks, which would be an important practical advantage. 

\begin{figure}[!h]
\begin{center}
\includegraphics[width=.7\linewidth]{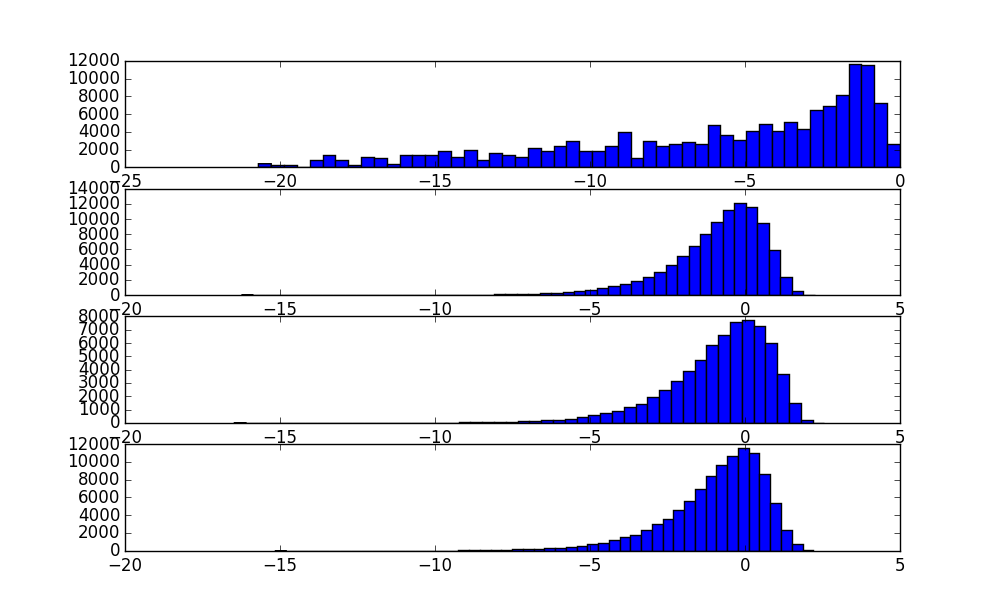}
\caption{Histogram of representations at each layer while training a fully connected network for MNIST. The figure represents a snap-shot in the middle of training. Each subfigure, from bottom up, represents the histogram of hidden states from the first layer to the last layer. The horizontal axes stand for the exponent of the layers' representations, i.e., $\log_{2}{\mathbf{x}}$. }
\label{repdist}
\end{center}
\end{figure}

The $\mathbf{x}$ to be quantized is not necessarily distributed symmetrically around $2$. For example, Figure~\ref{repdist} shows the distribution of $\mathbf{x}$ at each layer in the middle of training. The maximum amount of shift to the left does not need to be the same as that on the right. A more efficient way is to use different values for the maximum left shift and the maximum right shift. Bearing that in mind, we set it to $3$ bits maximum to the right and $4$ bits to the left.

%\subsection{Robustness over different optimizers}
%We repeated our performance evaluation using Adam [\cite{kingma2014adam}] to assess how robust the results are with respect to the choice of optimization procedure. Results are shown in Table \ref{adams}.

%NOTE: this table is fake for the moment, need to do experiments to fill into it.

%\begin{table}[!h]
%    \caption{Performances across all datasets}  \label{adams}
%    \begin{center}
%        \begin{tabular}{c | c | c | c | c}
%        \hline\hline
%	        & Full precision	& Binary connect
%	        & \begin{tabular}{c} Binary connect + \\ Quantized backprop
%	          \end{tabular}
%	        & \begin{tabular}{c} Ternary connect + \\ Quantized backprop
%	          \end{tabular}  \\
%        \hline
%        MNIST	& 1.33\%	& 1.23\%    & 1.29\%    & 1.15\%    \\
%        CIFAR10	& 15.64\%	& 12.04\%	& 12.08\%	& 12.01\%   \\
%        SVHN	& 2.85\%    & 2.47\%    & 2.48\%    & 2.42\%    \\
%        \hline\hline
%        \end{tabular}
%    \end{center}
%\end{table}

\section{Conclusion and future work}
We proposed a way to eliminate most of the floating point multiplications used during training a feedforward neural network. This could make it possible to dramatically accelerate the training of neural networks by using dedicated hardware implementations. 

A somewhat surprising fact is that instead of damaging prediction accuracy the approach tends improve it, which is probably due to several facts. First is the regularization effect that the stochastic sampling process entails. Noise injection brought by sampling the weight values can be viewed as a regularizer, and that improves the model generalization. The second fact is low precision weight values. Basically, the generalization error bounds for neural nets depend on the weights precision. Low precision prevents the optimizer from finding solutions that require a lot of precision, which correspond to very thin (high curvature) critical points, and these minima are more likely to correspond to overfitted solutions then broad minima (there are more functions that are compatible with such solutions, corresponding to a smaller description length and thus better generalization). Similarly, \cite{neelakantan2015adding} adds noise into gradients, which makes the optimizer prefer large-basin areas and forces it to find broad minima. It also lowers the training loss and improves generalization.

Directions for future work include exploring actual implementations of this approach (for example, using FPGA), seeking more efficient ways of binarization, and the extension to recurrent neural networks. 

\subsubsection*{Acknowledgments}
The authors would like to thank the developers of Theano~\citep{Bastien-Theano-2012}.
% optionally: Also, the authors thank * for insightful comments and discussion.
We acknowledge the support of the following agencies for research funding and
computing support: Samsung, NSERC, Calcul Qu\'{e}bec, Compute Canada,
the Canada Research Chairs and CIFAR.

\bibliography{qbp}
\bibliographystyle{iclr2016_conference}

\end{document}